\documentclass[twocolumn]{article}
\usepackage{graphicx}
\usepackage{amsmath}
\usepackage{amssymb}
\usepackage{times} 
\usepackage{hyperref} 
\usepackage{multirow}

\usepackage{booktabs}  
\usepackage{siunitx}  
\usepackage{array}   

\usepackage{xcolor}    

\usepackage{siunitx}  % 用于S列（数值对齐）
\usepackage{booktabs} % 用于\toprule/\midrule/\bottomrule

\hypersetup{
    colorlinks=false,    % 开启颜色链接（关闭则是边框样式）
    linkcolor=black,    % 内部链接（如章节引用）颜色
    urlcolor=black,     % URL链接颜色（重点配置）
    citecolor=black,    % 文献引用颜色（如果需要）
    filecolor=black,    % 文件链接颜色
    pdfborder={0 0 0},  % 取消超链接的边框（可选）
    pdfstartview=FitH   % PDF打开时的视图（可选）
}

\usepackage[ruled,vlined]{algorithm2e} 

\usepackage{tikz}
\usetikzlibrary{shapes.geometric, arrows.meta, positioning, fit, backgrounds, calc, patterns, decorations.pathreplacing}

\definecolor{moduleblue}{RGB}{70, 130, 180}
\definecolor{moduleorange}{RGB}{255, 140, 0}
\definecolor{modulegreen}{RGB}{60, 179, 113}
\definecolor{lightgray}{RGB}{240, 240, 240}
\definecolor{darkgray}{RGB}{100, 100, 100}

\tikzset{
    arrow/.style={-Stealth, thick},
    doublearrow/.style={Stealth-Stealth, thick},
    box/.style={rectangle, draw, thick, minimum width=2.5cm, minimum height=0.8cm, align=center},
    module/.style={rectangle, draw, thick, rounded corners, minimum width=3cm, minimum height=1.2cm, align=center, fill=white},
    stage/.style={rectangle, draw, thick, minimum width=2cm, minimum height=0.7cm, align=center, fill=white},
    comm/.style={rectangle, draw, thick, dashed, minimum width=2cm, minimum height=0.7cm, align=center, fill=red!10},
}

\title{\textbf{Accelerating Video Generation Inference with Sequential-Parallel 3D Positional Encoding Using a Global Time Index}}

\author{
    \textbf{Chao Yuan, Pan Li} \\
    \{\textit{storyiconx@gmail.com, pandalee@ustc.edu}\}  
}

\usepackage[style=numeric]{biblatex}
\usepackage{setspace}  % 加载间距控制包
\begin{document}
\maketitle

\begin{abstract}
Diffusion Transformer (DiT)-based video generation models inherently suffer from bottlenecks in long video synthesis and real-time inference, which can be attributed to the use of full spatiotemporal attention. Specifically, this mechanism leads to explosive \(O(N^2)\) memory consumption and high first-frame latency. To address these issues, we implement system-level inference optimizations for a causal autoregressive video generation pipeline. We adapt the Self-Forcing causal autoregressive framework to sequence parallel inference and implement a sequence-parallel variant of the causal rotary position embedding which we refer to as Causal-RoPE SP. This adaptation enables localized computation and reduces cross-rank communication in sequence parallel execution. In addition, computation and communication pipelines are optimized through operator fusion and RoPE precomputation. Experiments conducted on an eight GPU A800 cluster show that the optimized system achieves comparable generation quality, sub-second first-frame latency, and near real-time inference speed. For generating five second 480P videos, a 1.58\texttimes{} speedup is achieved, thereby providing effective support for real-time interactive applications.
\end{abstract}

\section{Introduction}
With the rapid advancement of deep learning, video generation technology has evolved from short clips to complex long sequences, driving diverse applications. Diffusion Transformer (DiT)-based models have become mainstream for high-quality video synthesis due to strong spatiotemporal consistency modeling capabilities. Building on the Transformer architecture\cite{NIPS2017_3f5ee243} and denoising diffusion probabilistic models\cite{NEURIPS2020_4c5bcfec}, recent video generation systems have achieved remarkable progress. For example, Wan2.1—an open-source large-scale video generation foundation model—adopts full spatiotemporal attention and Flow Matching as its training framework, achieving leading performance on multiple benchmarks\cite{wan2025wanopenadvancedlargescale}. However, the full-frame parallel inference design of such models severely restricts their scalability and practicality in real-world scenarios.

Three core bottlenecks hinder their application in long video generation and real-time inference. First, the $O(N^2)$ self-attention complexity\cite{10.1145/3530811} leads to quadratic memory growth with token count, making single-GPU long video inference impractical. Second, the fixed-length assumption of global parallel attention causes obvious temporal seams and degraded long-range consistency when generating videos exceeding the training frame limit. Third, bidirectional dependencies in global diffusion models prevent streaming inference—since current frame generation is affected by future frames, the system must wait for full video generation before output, leading to first-frame latency of tens of seconds or longer. These issues stem from the global diffusion paradigm and cannot be resolved by incremental engineering optimizations alone, requiring structural transformation.

Self-Forcing addresses these bottlenecks by transforming full-frame parallel diffusion models into causal autoregressive generators, enabling arbitrary-length video generation\cite{huang2025selfforcingbridgingtraintest}. Nevertheless, the official Self-Forcing codebase does not provide a production-ready sequence parallel implementation for multi GPU inference and its 3D RoPE computation relies on full sequence information. Both limitations induce significant cross rank communication overhead when scaling to multiple GPUs.

This paper presents system-level inference optimizations for Self-Forcing's causal autoregressive architecture. We implement a sequence-parallel baseline that builds on prior sequence-parallel designs such as Ulysses and on tensor and model parallel practices exemplified by Megatron CP. We develop three practical improvements. First, sequence-parallel integration adapted to causal KV cache workflows. Second, a sequence-parallel variant of the causal rotary position embedding which we call Causal-RoPE SP. Third, computation and communication pipeline optimizations including operator fusion and RoPE precomputation.

Experimental results on an eight GPU NVIDIA A800 cluster using bfloat16 precision show sub-second first-frame latency and a 36.97 percent end-to-end improvement corresponding to a 1.58 times speedup for five second 480P videos while preserving generation quality. This work offers an effective engineering path toward scalable low-latency long video inference.

\section{Related Work}
\subsection{Long Video Generation and Diffusion Model Inference}
Diffusion Transformer-based models, such as Wan2.1 \cite{wan2025wanopenadvancedlargescale}, have achieved state-of-the-art performance in high-quality short video synthesis. Built upon the Flow Matching training framework \cite{flowmatching,NEURIPS2018_69386f6b}, Wan2.1 employs full spatiotemporal attention for bidirectional information flow and global consistency modeling, coupled with a 3D Causal VAE\cite{kingma2022autoencodingvariationalbayes,wu2024improved} for spatiotemporal compression. However, the design of full spatiotemporal attention introduces significant computational bottlenecks: the $O(N^2)$ complexity leads to quadratic memory growth with sequence length, making long-video inference infeasible. To address this, Self-Forcing \cite{huang2025selfforcingbridgingtraintest} proposes an innovative solution by transforming parallel diffusion models into causal autoregressive generators, enabling video generation of arbitrary length through KV caching and a rolling window mechanism.

Nevertheless, Self-Forcing faces critical limitations in practical deployment. First, its reference implementation lacks production-ready support for Sequence Parallelism, restricting its scalability in multi-GPU environments. Second, its positional encoding computation still relies on global sequence information, which introduces significant communication overhead in distributed settings. These limitations hinder Self-Forcing from fully realizing its theoretical advantages for long-sequence generation. Our work targets these engineering shortcomings and proposes system-level optimizations.

\subsection{Sequence Parallelism}
Sequence Parallelism is a key strategy for scaling long-sequence models to multi-GPU environments. By partitioning the sequence dimension across GPU devices while keeping model parameters globally shared, it effectively mitigates the memory pressure induced by self-attention mechanisms \cite{li2021sequence}. Prior works, such as Ulysses \cite{jacobs2023deepspeed}, have demonstrated the effectiveness of SP for natural language processing Transformers. However, Ulysses is primarily designed to improve throughput through batch-level parallelism, whereas our work targets end-to-end latency reduction in autoregressive generation scenarios. Existing SP techniques remain under-explored for causal autoregressive video generation models. In particular, current parallelization schemes for video generation mostly target non-causal architectures and lack systematic support for KV caching and causal attention masking. Our work fills this gap by implementing a fully SP-compatible scheme for the Self-Forcing framework, addressing the unique challenges of causal autoregressive generation.

\subsection{Positional Encoding}
3D positional encoding is crucial for modeling spatiotemporal relationships among video tokens. Rotary Position Embedding (RoPE)\cite{su2023roformerenhancedtransformerrotary} has emerged as an effective approach for encoding positional information through rotation matrices. Wan2.1 employs a complex-number-based 3D RoPE \cite{arnab2021video}, while Qwen2.5-VL utilizes a $\cos$/$\sin$-based M-RoPE \cite{qwen2025qwen2}. These designs extend relative position encoding concepts\cite{shaw-etal-2018-self} to spatiotemporal domains. Existing schemes typically require access to the full sequence information to compute positional encodings, which introduces expensive cross-device communication in SP scenarios. Although Self-Forcing alleviates sequence length limitations through its autoregressive generation mechanism, its positional encoding computation still depends on global temporal information and fails to fully exploit the computational locality inherent in causal generation. Our optimization addresses this issue by proposing a positional encoding scheme that supports local computation, significantly reducing communication overhead in distributed environments.

\section{Methodology}
\subsection{Overall Optimization Framework}
Our work focuses on system-level inference optimizations for Self-Forcing's existing causal autoregressive architecture, without modifying its core causal reasoning logic . The optimization framework targets two critical limitations of the official Self-Forcing implementation: lack of Sequence Parallelism support and redundant communication caused by 3D RoPE in parallel scenarios.

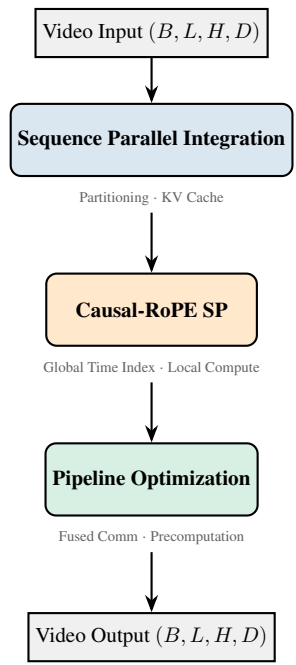
\begin{figure}[t]
    \centering
    \begin{tikzpicture}[node distance=0.6cm, scale=0.8, every node/.style={scale=0.8}]

        % Input
        \node[box, fill=lightgray, minimum width=3.5cm] (input) {Video Input $(B, L, H, D)$};

        % Module 1: Sequence Parallel
        \node[module, fill=moduleblue!20, minimum width=3.5cm, below=of input] (sp)
            {\textbf{Sequence Parallel Integration}};
        \node[font=\scriptsize, text=darkgray, below=0.1cm of sp] (sp_detail)
            {Partitioning $\cdot$ KV Cache};

        % Module 2: Causal-RoPE SP
        \node[module, fill=moduleorange!20, minimum width=3.5cm, below=0.8cm of sp_detail] (rope)
            {\textbf{Causal-RoPE SP}};
        \node[font=\scriptsize, text=darkgray, below=0.1cm of rope] (rope_detail)
            {Global Time Index $\cdot$ Local Compute};

        % Module 3: Pipeline Optimization
        \node[module, fill=modulegreen!20, minimum width=3.5cm, below=0.8cm of rope_detail] (pipeline)
            {\textbf{Pipeline Optimization}};
        \node[font=\scriptsize, text=darkgray, below=0.1cm of pipeline] (pipe_detail)
            {Fused Comm $\cdot$ Precomputation};

        % Output
        \node[box, fill=lightgray, minimum width=3.5cm, below=0.8cm of pipe_detail] (output)
            {Video Output $(B, L, H, D)$};

        % Arrows
        \draw[arrow] (input) -- (sp);
        \draw[arrow] (sp_detail) -- (rope);
        \draw[arrow] (rope_detail) -- (pipeline);
        \draw[arrow] (pipe_detail) -- (output);

    \end{tikzpicture}
    \caption{Overall optimization framework with three key modules.}
    \label{fig:system_architecture}
\end{figure}

The three key optimization modules are sequence-parallel integration, the Causal-RoPE SP design, and computation and communication pipeline optimization, as illustrated in Figure~\ref{fig:system_architecture}. These modules work together as follows. First, sequence-parallel partitioning splits the sequence dimension across GPU ranks to reduce single GPU memory pressure. Second, Causal-RoPE SP encodes global temporal position per block so RoPE can be computed locally and cross-rank communication can be reduced. Third, operator fusion and RoPE precomputation reduce kernel launch overhead and host GPU interactions.

% ========================================
% Figure: Overall System Architecture
% ========================================

% ========================================
% Figure: Pipeline Comparison (Baseline vs Optimized)
% ========================================
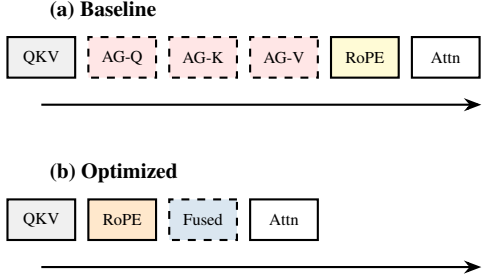
\begin{figure}[t]
    \centering
    \begin{tikzpicture}[
        node distance=0.15cm,
        scale=0.9,
        every node/.style={scale=0.9},
        smallstage/.style={rectangle, draw, thick, minimum width=1cm, minimum height=0.6cm, align=center, fill=white},
        smallcomm/.style={rectangle, draw, thick, dashed, minimum width=1cm, minimum height=0.6cm, align=center, fill=red!10}
    ]

        % Baseline Pipeline (Top)
        \node[font=\small\bfseries, anchor=west] (baseline_title) at (0, 3.2) {(a) Baseline};

        \node[smallstage, fill=lightgray] (b1) at (0, 2.5) {\scriptsize QKV};
        \node[smallcomm, right=of b1] (b2) {\scriptsize AG-Q};
        \node[smallcomm, right=of b2] (b3) {\scriptsize AG-K};
        \node[smallcomm, right=of b3] (b4) {\scriptsize AG-V};
        \node[smallstage, fill=yellow!20, right=of b4] (b5) {\scriptsize RoPE};
        \node[smallstage, right=of b5] (b6) {\scriptsize Attn};

        % Timeline arrow
        \draw[-Stealth, thick] (0, 1.8) -- (6.5, 1.8);
        \node[font=\tiny, text=red, above] at (3.5, 2.8) {};

        % Optimized Pipeline (Bottom)
        \node[font=\small\bfseries, anchor=west] (opt_title) at (0, 0.8) {(b) Optimized};

        \node[smallstage, fill=lightgray] (o1) at (0, 0.1) {\scriptsize QKV};
        \node[smallstage, fill=moduleorange!20, right=of o1] (o2) {\scriptsize RoPE};
        \node[smallcomm, fill=moduleblue!20, right=of o2] (o3) {\scriptsize Fused};
        \node[smallstage, right=of o3] (o4) {\scriptsize Attn};

        % Timeline arrow
        \draw[-Stealth, thick] (0, -0.6) -- (6.5, -0.6);

    \end{tikzpicture}
    \caption{Pipeline comparison. (a) Baseline: sequential AllGather then RoPE. (b) Optimized: local RoPE with fused communication (10.1× faster).}
    \label{fig:pipeline_comparison}
\end{figure}

\subsection{Sequence Parallel Implementation}
To overcome the lack of Sequence Parallelism support in Self-Forcing, we implement a full SP solution adapted to its causal autoregressive properties and KV caching mechanism. Unlike Ulysses, which targets throughput improvement through batch-parallel workloads, our implementation focuses on minimizing end-to-end latency for streaming video generation.

Our SP implementation partitions the sequence dimension evenly across $P$ parallel GPU ranks, with each rank holding only a local subsequence of length $L/P$. This design effectively reduces the memory footprint on individual GPUs, enabling the processing of longer video sequences. The key challenge in this implementation is maintaining causal attention consistency and ensuring compatibility with the KV caching mechanism\cite{10.1145/3600006.3613165} under sequence partitioning. Causal attention requires that a current token can only attend to previous tokens, a property that must be preserved across rank boundaries. Simultaneously, the KV cache must be correctly updated and shared among ranks to avoid redundant computations and ensure consistency across the distributed system.

We first develop a baseline SP implementation (Algorithm~1) where each GPU rank projects its local subsequence to Q, K, V tensors, aggregates them across ranks for full-context attention, then redistributes the output to restore sequence partitioning.

However, this baseline implementation suffers from two critical inefficiencies that hinder its performance. First, it requires three separate AllGather operations—one each for the Q, K, and V tensors—to gather the full sequence across all ranks, introducing significant communication overhead. Second, the 3D RoPE computation depends on the complete sequence information, forcing it to wait for the completion of the AllGather operations. This sequential dependency between communication and computation prevents overlapping, resulting in idle GPU cycles and reduced overall throughput, as illustrated in Figure~\ref{fig:pipeline_comparison}. These limitations motivate the optimizations proposed in subsequent sections to enhance the efficiency of the SP implementation within the Self-Forcing framework.

\begin{algorithm}[htbp]
\caption{Sequence Parallel Self Attention Baseline}
\KwIn{
Input $x \in \mathbb{R}^{B \times \frac{L}{P} \times H \times D}$; \\
Sequence parallel degree $P$; \\
Grid sizes $(F, H, W)$; \\
RoPE frequencies $\theta$; \\
Global start frame offset $s$
}
\KwOut{
Output $o \in \mathbb{R}^{B \times \frac{L}{P} \times H \times D}$
}

\BlankLine
\textbf{Step 1: Local QKV projection}\;
$q, k, v \gets \mathrm{QKV}(x)$

\BlankLine
\textbf{Step 2: Gather full sequence across sequence-parallel ranks}\;
$q \gets \mathrm{AllGather}(q, \mathrm{dim}=1)$\;
$k \gets \mathrm{AllGather}(k, \mathrm{dim}=1)$\;
$v \gets \mathrm{AllGather}(v, \mathrm{dim}=1)$

\BlankLine
\textbf{Step 3: Apply 3D RoPE with global temporal indexing}\;
$\tilde{q}, \tilde{k} \gets \mathrm{RoPE}(q, k, (F, H, W), \theta, s)$

\BlankLine
\textbf{Step 4: Split head dimension for tensor parallelism}\;
$\tilde{k} \gets \mathrm{Split}(\tilde{k}, \mathrm{dim}=2)$\;
$v \gets \mathrm{Split}(v, \mathrm{dim}=2)$

\BlankLine
\textbf{Step 5: Update KV cache}\;
$\mathrm{KVCache}[\text{key}] \gets \tilde{k}$\;
$\mathrm{KVCache}[\text{value}] \gets v$

\BlankLine
\textbf{Step 6: Attention computation}\;
$\tilde{q} \gets \mathrm{Split}(\tilde{q}, \mathrm{dim}=2)$\;
$k \gets \mathrm{KVCache}[\text{key}]$\;
$v \gets \mathrm{KVCache}[\text{value}]$\;
$o \gets \mathrm{Attention}(\tilde{q}, k, v)$

\BlankLine
\textbf{Step 7: All-to-all exchange to restore output layout}\;
$o \gets \mathrm{AllToAll}(o, \mathrm{scatter\_dim}=1, \mathrm{gather\_dim}=2)$

\Return{$o$}

\end{algorithm}

\subsection{Causal-RoPE SP: Positional Encoding for Block-Wise Autoregression}
To resolve the communication inefficiency of traditional 3D RoPE in sequence-parallel mode, we adopt the causal RoPE design used in Self-Forcing and implement a sequence-parallel variant which we call Causal-RoPE SP. The Causal-RoPE SP design integrates global temporal indices so that RoPE computation can be performed locally on each rank with reduced cross-rank communication.

\subsubsection{Background: 3D RoPE in Video Generation}
Wan2.1 adopts 3D RoPE to encode the spatiotemporal positions of video tokens, splitting rotation frequencies into temporal ($\theta_T$), height ($\theta_H$), and width ($\theta_W$) dimensions. Its design is analogous to Qwen2.5-VL's M-RoPE but differs in implementation:
\begin{itemize}
    \item Qwen2.5-VL uses a $\cos/\sin$ form with $\text{rotate\_half}$ ;
    \item Wan2.1 uses explicit complex-number multiplication to model rotation , which aligns with RoPE's mathematical essence:
    \[
    \text{RoPE}(x, \theta) = x \cdot e^{i\theta} = x \cdot (\cos\theta + i\sin\theta)
    \]
\end{itemize}

\subsubsection{Design of Causal-RoPE SP}

\begin{figure}[t]
    \centering
    \begin{tikzpicture}[node distance=0.5cm, scale=1.10, every node/.style={scale=1.10}]

        % Title label
        \node[font=\scriptsize\bfseries] at (2.2, 3.2) {Sequence Partitioning Across Ranks};

        % Sequence partitioning across ranks
        \node[draw, thick, fill=moduleblue!20, minimum width=1.1cm, minimum height=0.6cm] (r0) at (0, 2.5) {\scriptsize Rank 0};
        \node[draw, thick, fill=moduleblue!30, minimum width=1.1cm, minimum height=0.6cm, right=0.08cm of r0] (r1) {\scriptsize Rank 1};
        \node[draw, thick, fill=moduleblue!40, minimum width=1.1cm, minimum height=0.6cm, right=0.08cm of r1] (r2) {\scriptsize Rank 2};
        \node[draw, thick, fill=moduleblue!50, minimum width=1.1cm, minimum height=0.6cm, right=0.08cm of r2] (r3) {\scriptsize Rank 3};

        \draw[decorate, decoration={brace, amplitude=3pt, mirror}]
            (r0.south west) -- (r3.south east)
            node[midway, below=2pt, font=\scriptsize] {Full Sequence $L$};

        % Highlight Rank 1 with zoom
        \draw[thick, moduleorange, rounded corners] ($(r1.south west)+(-0.05,-0.02)$) rectangle ($(r1.north east)+(0.05,0.02)$);

        % Arrow pointing down to detail
        \draw[-Stealth, thick, moduleorange] (r1.south) -- ++(0, -0.6);

        % Detail box - directly below Rank 1, aligned with sequence
        \node[draw, thick, rounded corners, fill=moduleorange!5, minimum width=4.5cm, minimum height=1.8cm]
            (detail_box) at (2.2, 0.5) {};

        \node[font=\scriptsize\bfseries, anchor=north] at (detail_box.north) {Token Index Computation};

        \node[align=left, font=\scriptsize, anchor=center] at (2.2, 0.3) {
            $i_{\text{global}} = r \cdot \frac{L}{P} + i_{\text{local}}$ \\[3pt]
            $t_{\text{global}} = s + \lfloor i_{\text{global}}/(H \cdot W) \rfloor$
        };

        % Result box - RoPE formula
        \draw[-Stealth, thick] (detail_box.south) -- ++(0, -0.4);

        \node[draw, thick, fill=modulegreen!15, rounded corners, align=center, font=\scriptsize, minimum width=4.5cm]
            (result_box) at (2.2, -1.4) {
            $\text{RoPE}(x) = x \odot e^{i(t_{\text{global}}\theta_T + h\theta_H + w\theta_W)}$ \\[2pt]
            \textbf{Local computation, no communication}
        };

    \end{tikzpicture}
    \caption{Causal-RoPE SP: local computation of global time index via start\_frame parameter $s$.}
    \label{fig:causal_rope_sp}
\end{figure}
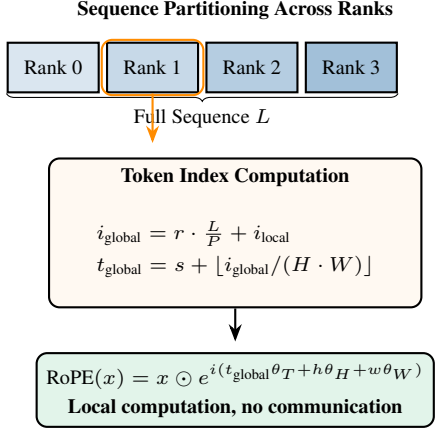

The original Self-Forcing RoPE implementation computes positional encodings after gathering the full sequence via AllGather, creating a sequential dependency that prevents computation-communication overlap. To address this, we leverage the $\text{start\_frame}$ parameter that Self-Forcing uses to encode the global temporal position of each generation block, enabling each rank to compute its global temporal indices locally. The key modifications are:
\begin{enumerate}
    \item Global Time Index: For a token at local spatiotemporal position t, h, w in block k with block size $\tau$, where t denotes the local time index within the current block, its global time index is $t_{\text{global}} = t + s$ where s equals k times $\tau$ and start\_frame equals s.
    As an example, with chunk wise generation using three latent frames per block so that $\tau$ equals three, the start\_frame values for successive generation blocks are 0, 3, 6 and so on.
    \item Localized Computation: The encoding formula is:
    \[
    x'_{t,h,w} = x_{t,h,w} \odot e^{i(t_{\text{global}}\theta_T + h\theta_H + w\theta_W)}
    \]
    \item Algorithm Implementation: Figure~\ref{fig:causal_rope_sp} illustrates this design. As shown in Algorithm 2, Causal-RoPE SP computes the global time index for each local token independently, without requiring full sequence information.
\end{enumerate}

\begin{algorithm}[htbp]
\caption{Causal-RoPE SP Implementation}
\KwIn{Input tensor $x$; Grid sizes $(F, H, W)$; RoPE frequencies $\text{freqs}$; Global start frame $s$}
\KwOut{Rotated tensor $x_{\text{causal\_rot}}$}
$\text{freqs}_i = \text{torch.cat}([$
    $\text{freqs}[0][s:s+F].view(F, 1, 1, -1).expand(F, H, W, -1),$
    $\text{freqs}[1][:H].view(1, H, 1, -1).expand(F, H, W, -1),$
    $\text{freqs}[2][:W].view(1, 1, W, -1).expand(F, H, W, -1)$
$], \text{dim}=-1)$\;
$x_i = \text{torch.view\_as\_complex}(x[i, :\text{seq\_len}].to(\text{torch.float64}).reshape(\text{seq\_len}, n, -1, 2))$\;
$x_i = \text{torch.view\_as\_real}(x_i \cdot \text{freqs}_i).flatten(2)$\;
$x_{\text{causal\_rot}} = x_i$\;
\Return{$x_{\text{causal\_rot}}$}
\end{algorithm}

% ========================================
% Figure: Causal-RoPE SP Working Principle
% ========================================

\subsubsection{Feasibility of Local Computation}
For a local token with index $i_{\text{local}}$ on rank $r$ (where each rank handles the interval $[r \cdot L/P, (r+1) \cdot L/P)$):
\begin{itemize}
    \item Its global sequence position is $i_{\text{global}} = r \cdot L_{\text{local}} + i_{\text{local}}$;
    \item Its local time index within the current block can be derived as $t = \lfloor i_{\text{global}}/(H \times W) \rfloor$;
    \item Its global time index is then computed as $t_{\text{global}} = s + \lfloor i_{\text{global}}/(H \times W) \rfloor$ ( $s = \text{start\_frame}$ ).
\end{itemize}

This formulation is equivalent to the compact expression $t_{\text{global}} = t + s$ presented in Section 3.3.2, where $t$ represents the derived local time index. Since $t_{\text{global}}$ can be computed using only local rank information, local token index and shared parameters $H, W, s$, Causal-RoPE SP achieves full localized computation without cross-rank communication while preserving causal consistency and global temporal order.

\subsection{Overlapped Computation and Communication}
Building on sequence-parallel integration and Causal-RoPE SP, we further optimize the pipeline by fusing communication operations and precomputing RoPE frequencies, eliminating redundant overhead.

\subsubsection{Key Optimizations}
\begin{enumerate}
    \item Fused All-to-All Communication: Replaces the three separate AllGather and one Split operations in the baseline with a single fused FusedAllToAll (Algorithm 3) \cite{punniyamurthy2024optimizingdistributedmlcommunication}, which simultaneously gathers the sequence dimension and splits the attention head dimension. This reduces communication rounds and data movement.
    \item RoPE Frequency Precomputation: Replaces dynamic LRU caching of $\cos/\sin$ frequencies with precomputation and storage in continuous tensors. This bypasses Host-GPU communication (Host Op) and enables direct GPU addressing during inference.
    \item Operator Fusion with TileLang: Fuses QKV projection and Causal-RoPE computation into a single kernel, reducing kernel launch overhead and improving data locality\cite{tilelang2025,222575}. Compared with the commonly used Triton implementation in the community, this fusion achieves an approximately 10\% performance improvement.
\end{enumerate}

\begin{algorithm}[htbp]
\caption{Sequence-Parallel Self-Attention (Optimized)}
\KwIn{
Input $x \in \mathbb{R}^{B \times \frac{L}{P} \times H \times D}$; \\
Sequence parallel degree $P$; \\
Sequence-parallel rank $r$; \\
Grid sizes $(F, H, W)$; \\
RoPE frequencies $\theta$; \\
Global start frame offset $s$
}
\KwOut{
Output $o \in \mathbb{R}^{B \times \frac{L}{P} \times H \times D}$
}

\BlankLine
\textbf{Step 1: Local QKV projection}\;
$q, k, v \gets \mathrm{QKV}(x)$

\BlankLine
\textbf{Step 2: Compute global token offset}\;
$L_{\text{local}} \gets \frac{L}{P}$\;
$\Delta \gets r \cdot L_{\text{local}}$\;

\textbf{Step 3: Vectorized Causal 3D RoPE (no communication)}\;
$\tilde{q} \gets \mathrm{CausalRoPE}(q, (F, H, W), \theta, s, \Delta)$\;
$\tilde{k} \gets \mathrm{CausalRoPE}(k, (F, H, W), \theta, s, \Delta)$\;

\tcp{Internal computation: $I_{\text{global}} \gets \mathrm{arange}(\Delta, \Delta + L_{\text{local}})$, $T_{\text{global}} \gets s + \lfloor I_{\text{global}} / (H \cdot W) \rfloor$}

\BlankLine
\textbf{Step 4: Fused all-to-all (sequence gather + head split)}\;
$q, k, v \gets \mathrm{FusedAllToAll}(
\tilde{q}, \tilde{k}, v,\;
\mathrm{scatter\_dim}=2,\;
\mathrm{gather\_dim}=1
)$

\BlankLine
\textbf{Step 5: Update KV cache}\;
$\mathrm{KVCache}[\text{key}] \gets k$\;
$\mathrm{KVCache}[\text{value}] \gets v$

\BlankLine
\textbf{Step 6: Attention computation}\;
$k \gets \mathrm{KVCache}[\text{key}]$\;
$v \gets \mathrm{KVCache}[\text{value}]$\;
$o \gets \mathrm{Attention}(q, k, v)$

\BlankLine
\textbf{Step 7: All-to-all exchange to restore output layout}\;
$o \gets \mathrm{AllToAll}(o,\;
\mathrm{scatter\_dim}=1,\;
\mathrm{gather\_dim}=2)$

\Return{$o$}

\end{algorithm}

\subsubsection{Performance Comparison}
\begin{figure}[htbp]
    \centering
    \includegraphics[width=0.5\textwidth]{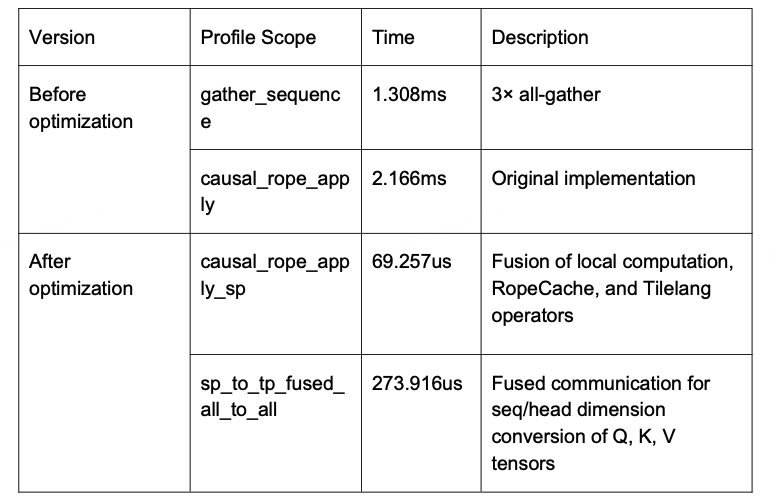}
    \caption{Profiling Results}
    \label{fig:profile_results}
\end{figure}

To quantify module-level improvements, we profile key operations before and after optimization. As shown in Figure~\ref{fig:profile_results}, the optimized pipeline reduces the combined latency of sequence gathering and RoPE computation from 3.474ms to 0.343ms per self-attention call. 

In a typical 5-second 480P video inference with the Self-Forcing framework, the system performs 920 self-attention computations in total (across all DiT blocks and denoising steps). Therefore, the end-to-end latency reduction from this module-level optimization is:
\[
\Delta T_{\text{e2e}} = 920 \times \Delta t_{\text{per-call}} \approx 2.88\,\text{s},
\]
where $\Delta t_{\text{per-call}} = (1.308 + 2.166) - (0.069257 + 0.273916) = 3.474 - 0.343 = 3.131\,\text{ms}$.

Compared to the baseline end-to-end latency of 8.86s, our optimizations achieve a 36.97\% speedup (1.58×) with no loss in generation quality , validating the effectiveness of the pipeline design.

\section{Experiments}
To validate the effectiveness of our proposed system-level inference optimizations for the Self-Forcing framework, we conduct comprehensive experiments focusing on inference performance, generation quality, and scalability. All experiments are performed on an 8-card NVIDIA A800 GPU cluster to ensure consistency with practical deployment scenarios for long video inference.

\subsection{Experimental Setup}
Experiments are conducted on an 8×A800 GPU cluster with PyTorch 2.6.0, CUDA 12.2, and NCCL 2.28.4. We evaluate 5-second 480P (832×480) video generation at 16 FPS, measuring end-to-end latency, first-frame latency, and throughput. The baseline uses Self-Forcing with basic SP (Algorithm 1) and AllGather for RoPE.

\subsection{Performance Evaluation}
We compare the inference performance of our optimized Self-Forcing  with the baseline on the 5-second 480P video generation task, focusing on key performance indicators and module-level latency breakdown.

\subsubsection{End-to-End Inference Performance and Resolution Scaling}
\begin{figure}[htbp]
    \centering
    \includegraphics[width=0.5\textwidth]{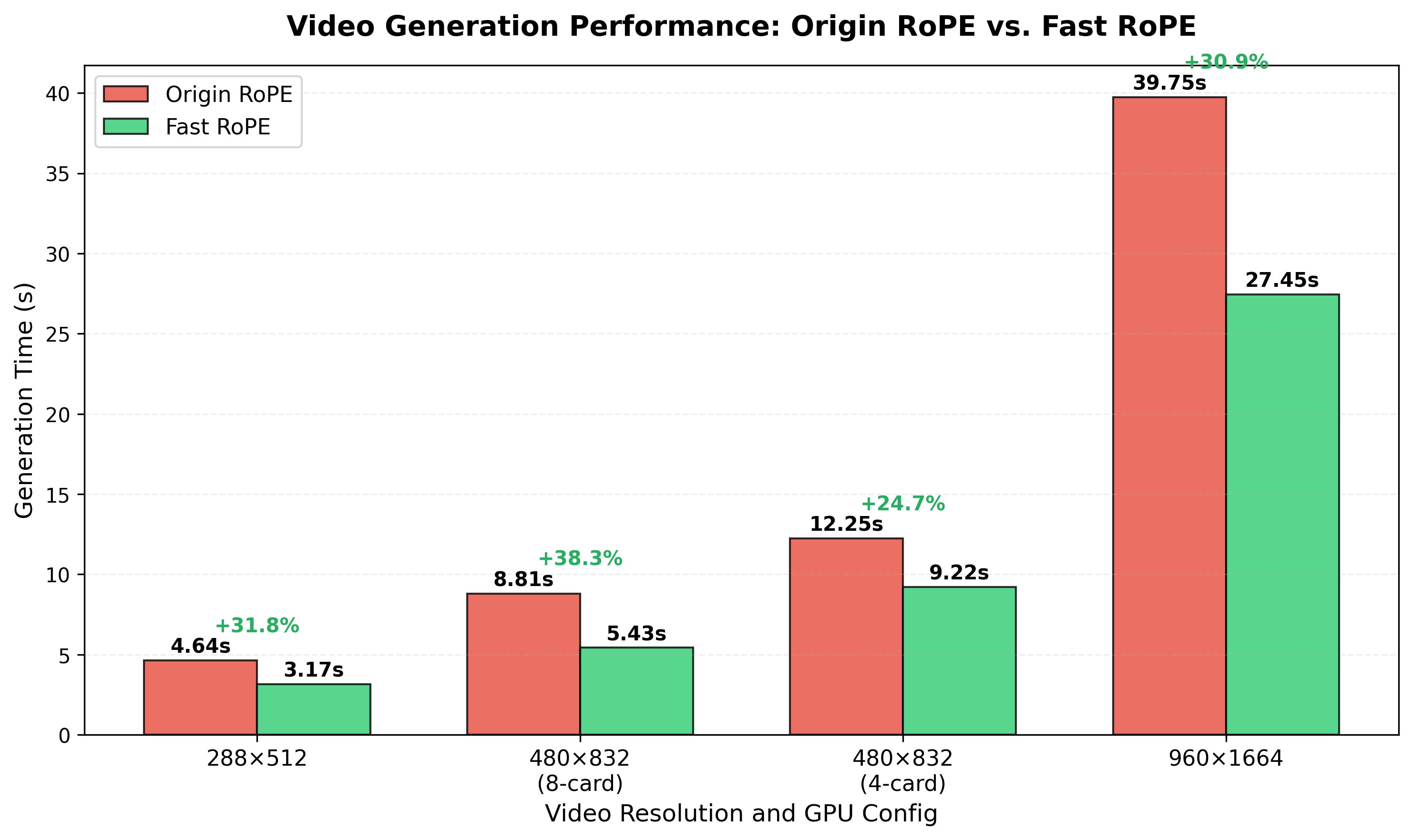}
    \caption{Consistent speedup across multiple resolutions and GPUs.}
    \label{fig:comparison_rope}
\end{figure}

To evaluate scalability across varying computational loads and parallelism degrees, we extend the evaluation to three resolutions (288×512, 480×832, 960×1664) on both eight GPU and four GPU configurations. Figure~\ref{fig:comparison_rope} demonstrates that the optimization maintains consistent effectiveness across different scales. On the eight GPU cluster, speedup ratios range from 1.46× to 1.62× across all resolutions, with the 480×832 resolution achieving the highest acceleration (8.81s → 5.43s). The four GPU configuration on 480×832 resolution yields a 1.33× speedup (12.25s → 9.22s), validating the optimization's robustness under reduced parallelism degrees while maintaining efficient resource utilization.

\subsubsection{Ablation Study}
\begin{figure}[htbp]
    \centering
    \includegraphics[width=0.5\textwidth]{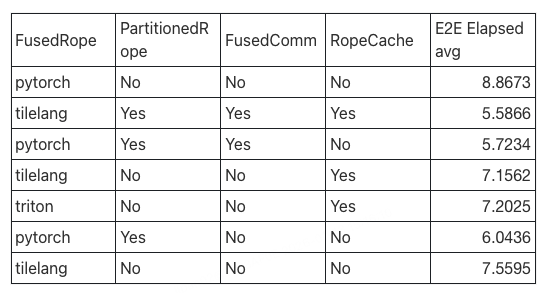}
    \caption{Ablation study.}
    \label{fig:ablation}
\end{figure}

We ran a set of ablation experiments to quantify the individual and combined contributions of the key optimization components. Figure~\ref{fig:ablation} lists per-configuration end-to-end elapsed time averaged over the measurement runs for the 5s 480P generation task.

For the 5-second 480P video task, this module-level optimization contributes approximately 2.88s to the total end-to-end latency reduction, which aligns with our theoretical analysis in Section~3.4.

\subsection{Experimental Summary}
Our experiments demonstrate that:
\begin{enumerate}
    \item Performance: The proposed optimizations achieve a 36.97\% speedup (1.58×) on 5-second 480P video generation, with sub-second first-frame latency;
    \item Scalability: SP support and memory-efficient design significantly extend the maximum supported video length, addressing the core bottleneck of long video inference.
\end{enumerate}

\section{Conclusion}
This paper presents a series of system-level practices for the inference optimization of video generation models, focusing on the computation and communication optimization of block-wise autoregressive video models in sequence parallel scenarios. In addition to the above work, we are also continuously exploring directions such as dynamic low-bit quantization and computation graph-level optimization, laying the foundation for subsequent larger-scale and lower-latency video generation systems.

\printbibliography

\end{document}